\documentclass{article}
\usepackage{spconf,amsmath,graphicx}

\usepackage{enumitem}
\setlist{nosep, leftmargin=14pt}
\usepackage{subfig}
\usepackage{mwe} % to get dummy images

\title{Uncertainty-guided Contrastive Learning for Single Source Domain Generalisation}
\name{Anastasios Arsenos $^1$,    Dimitrios Kollias $^2$, Evangelos Petrongonas,  Christos Skliros $^3$,  Stefanos Kollias $^1$ \vspace{0.3cm}}
\address{$^1$ School of Electrical \& Computer Engineering, National Technical University of Athens, Greece \\
$^{2}$  School of Electronic Engineering \& Computer Science, Queen Mary University of London, UK \\
$^3$ Hellenic Drones S.A., Greece}

\begin{document}
%\ninept
%
\maketitle

\begin{abstract}
In the context of single domain generalisation, the objective is for models that have been exclusively trained on data from a single domain to demonstrate strong performance when confronted with various unfamiliar domains. In this paper, we introduce a novel model referred to as Contrastive Uncertainty Domain Generalisation Network (CUDGNet). %This model is designed to address the challenge of single domain generalisation in the realm of image recognition. 
The key idea is to augment the source capacity in both input and label spaces through the fictitious domain generator and jointly learn the domain invariant representation of each class through contrastive learning.
%Our method is designed to achieve large domain expansion from the generator subnetwork while simultaneously avoiding representation collapse. 
Extensive experiments on two Single Source Domain Generalisation (SSDG) datasets demonstrate the effectiveness of our approach, which surpasses the state-of-the-art single-DG methods by up to $7.08\%$. Our method also provides efficient uncertainty estimation at inference time from a single forward pass through the generator subnetwork.
\end{abstract}
\section{Introduction}
\label{sec:intro}

The concept of leveraging diversity for model training has been extensively explored. Previous research \cite{pden, learningtodiversify, jigen} demonstrates that employing a wide array of augmentations during training enhances a model's resilience against shifts in distribution. When the nature of diversity encountered during testing is identifiable, particular augmentations can be applied.  

%Domain invariant representation learning (contrastive learning)
Apart from input diversity, SSDG approaches need to learn domain invariant representations of the data. Many previous works \cite{learningtodiversify, pden} incorporated contrastive learning to acquire domain-invariant representations. These representations ensure that each class forms distinct clusters, thereby enabling the learning of an improved decision boundary that enhances generalisation capabilities.
Nevertheless, prior research \cite{metacnn, mada, randconv} disregard the potential hazards associated with utilizing augmented data to address out-of-domain generalisation. This omission gives rise to significant apprehensions related to safety and security, particularly in contexts involving mission-critical applications. For example, in scenarios involving the deployment of self-driving vehicles in unfamiliar surroundings, a comprehensive understanding of predictive uncertainty becomes pivotal for effective risk evaluation.
%Adversarial data augmentation

%Style transfer
%Limitations (no uncertainty assessment)

Recently, \cite{uncertainty} proposed a Bayesian meta-learning framework that leverages the uncertainty of domain augmentations to improve the domain generalisation in a curriculum learning scheme and provides fast uncertainty assessment. 
 
Its main limitations included sensitivity to hyperparameters that made the training unstable and high computational demands that made it challenging to scale to complex networks. 

Inspired by \cite{uncertainty}, we aim to leverage the uncertainty of domain augmentations in both input and label spaces. In order to tackle the aforementioned limitations of previous works, we propose a novel framework that contains a 
task model $M$  and a domain augmentation generator $G$. These components enhance each other through collaborative learning. The domain augmentation generator produces secure and efficient domains, guided by uncertainty assessment. These generated domains are systematically extended to enhance coverage and comprehensiveness. To achieve cross-domain invariant representation across all generated domains, contrastive learning is introduced in the learning process of the task model $M$.

The main contributions can be summarized below:

\begin{figure*}[h!]
\centering
\includegraphics[width=.65\linewidth]{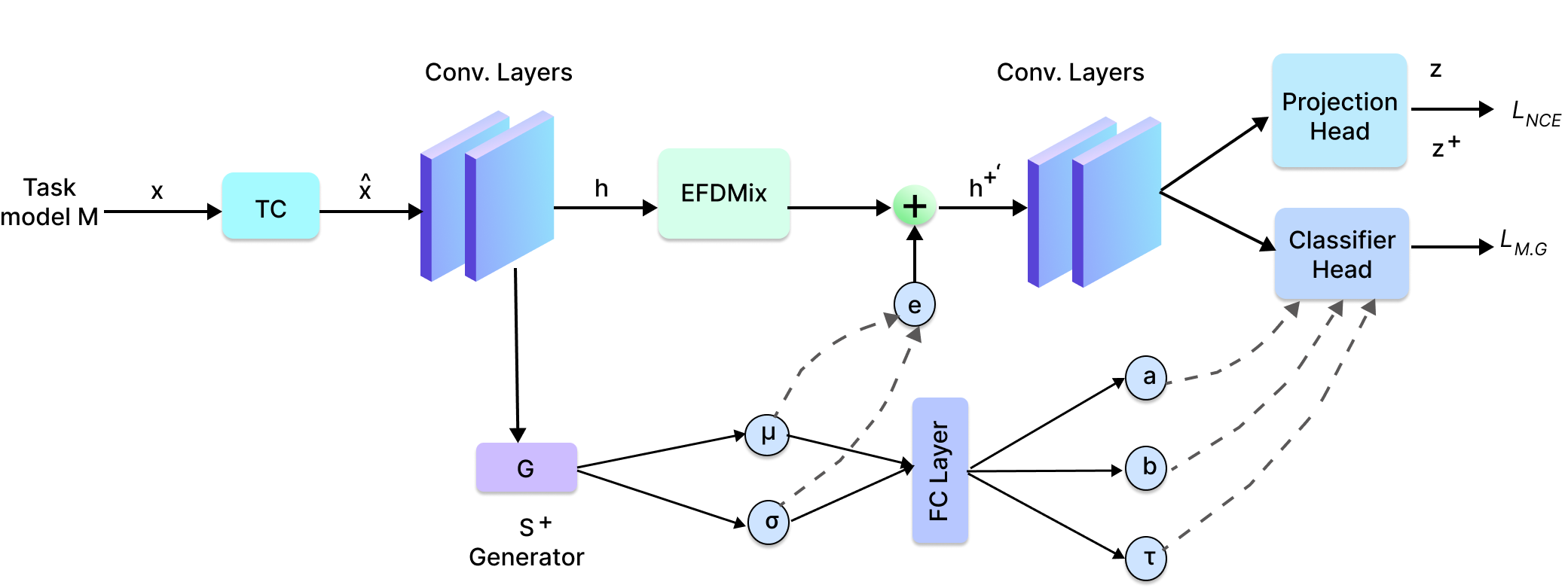}
\caption{The overall framework of the proposed CUDGNet. The Task Model M and the domain augmentation Generator G are jointly trained while the transformation component TC and style mixing (EFDMix) further enrich the augmentation capacity. The contrastive loss will guide semantically similar samples from different domains to be closer in the embedding space.}
\label{fig:model}
\end{figure*}

\begin{itemize}[noitemsep]
    \item We propose a novel framework that leverages adversarial data augmentation and style transfer for domain expansion while ensuring semantic information perseverance through contrastive learning. 
    \item  Our framework can estimate the uncertainty in a single forward pass while achieving state-of-the-art accuracy.
    \item We validate our framework's performance via comparison and ablation study on two SSDG datasets.
\end{itemize}

\vspace{-0.4cm}

\section{Materials and Method}
\label{sec:method}

Here, we present an outline of CUDGNet, as depicted in Figure \ref{fig:model}. Our objective involves training a robust model from a solitary domain $S$, with the anticipation that the model's effectiveness will extend across an unfamiliar domain distribution $\{T_1, T_2, ...\} \sim p(T)$. In order to achieve that, we create a sequence of domain augmentations $\{S_{1}^{+}, S_{2}^{+}, ...\} \sim p(S^{+})$ to closely approximate $p(T)$, allowing the task model to acquire the ability to generalise across domains that it has not encountered before. In addition, we show how we can assess the uncertainty of new domains as a byproduct of the perturbations used for adversarial domain augmentation. 

To generate new domains while retaining class-specific details we introduce two auxiliary components, namely the Transformation Component and the domain augmentation generator $G$ component. The latter is a novel feature perturbation subnetwork that combines style transfer and variational feature perturbations that follow a learnable multivariate Gaussian distribution $\mathcal{N}(\mu, \sigma^2)$, for diverse and content-preserved feature augmentations. 
%These perturbations explicitly capture the uncertainty related to the task model $M$ and utilize it to create the fictitious domains $\{S_{1}^{+}, S_{2}^{+}, ...\}$ by enhancing the capacity in both the input and label spaces. 
Additionally, the domain augmentation learning procedure is enriched by incorporating image-structure information generated through our image transformation component which utilizes both affine transformations and fractals to enrich the input space. To effectively organise domain alignment and classification, we incorporate contrastive learning to acquire representations that are invariant to domain shifts as well as to avoid representation collapse caused by extreme domain shifts of feature perturbations. This process facilitates the progressive formation of domain augmentations and well-defined clusters for each class in the representation space.

\vspace{0.15cm}

\noindent
\underline{\textbf{2.1. Transformation Component (TC)}}: it can transform the initial image $x$ from the original domain $\mathcal{S}$ into a novel image $\hat{x}$ within the same domain using the following process:

\vspace{-0.4cm}

\begin{equation}
\label{eq:tc}
  \hat{x} =   TC(x) = \begin{cases}
        (x \oplus f_1)  \\
        (x \oplus f_1) \otimes (x \oplus x_{aff})  \\
        (x \oplus f_1) \otimes (x \oplus x_{aff}) \otimes (x \oplus f_2) 
    \end{cases}  
\end{equation}

\noindent where: each of the three branches has an equal probability of occurring; $x_{aff}$ is the resulting image after an affine transformation (like rotation, translation and contrast) is applied to the initial image $x$; %Inspired by \cite{pixmix} we also include
$f_1$, $f_2$  denote fractal images \cite{fractals}; $\oplus$ and $\otimes$ are element-wise additions and multiplications respectively.

The transformation component may not expand the domain space but is a crucial part of our method. It helps to avoid representational collapse in the initial epochs of domain expansion where $G$ may induce severe noise and thus the whole framework becomes more robust.  Fractals exhibit specific structural characteristics, which are notably non-random and improbable to emerge from processes of maximum entropy or gaussian noise (perturbations); thus this transformation is orthogonal to the domain augmentation
generator $G$.  In Eq. \ref{eq:tc} the image is passed through $k$ transformations (with $k \in [0, 10]$ being a hyperparameter).

\vspace{0.15cm}

\noindent
\underline{\textbf{2.2. Domain augmentation
generator}}:
We illustrate the process of producing the unseen domain $S^{+}$ from $S$ through the generator $G$, ensuring that the samples generated adhere to the criteria of safety and effectiveness. Safety denotes that the generated samples preserve semantic information, while effectiveness implies that the generated samples encompass a diverse range of unseen domain-specific details.

\textbf{Style manipulation}. In conjunction with the Transformation Component, we enrich the input space using style manipulation. We integrate Exact Histogram Matching (EHM) \cite{ehm1} that can represent style information by using high-order feature statistics. While there are several strategies available for EHM, we choose to utilize the Sort-Matching algorithm \cite{sortmatch} because of its efficient execution speed. Sort-Matching is implemented by matching two sorted vectors, whose indexes are illustrated in a one-line notation: 

\vspace{-0.4cm}

\begin{equation}
\label{eq:efdm}
\begin{split}
    \textbf{w}: \tau = (\tau_{1} \quad \tau_{2} \quad \tau_{3} \quad ... \quad \tau_{n})   \\
    \textbf{r}: \kappa = (\kappa_{1} \quad\kappa_{2} \quad \kappa_{3} \quad ... \quad \kappa_{n})
\end{split}
\end{equation}

\noindent The Sort-Matching output is: $out_{\tau_{i}} = r_{k_{i}}$ \hspace{0.2cm} (3)

%\vspace{-0.1cm}

%\begin{equation}
%\label{eq:efdm2}
%    out_{\tau_{i}} = r_{k_{i}}
%\end{equation}

To create a wide range of feature augmentations that combine various styles, we rely on the Exact feature Distribution  mixing -outlined in Eq. \ref{eq:efdm3}- by incorporating interpolated sorted vectors. This leads to the development of the Exact Feature Distribution Mixing (EFDMix) method \cite{efdm}.

\vspace{-0.4cm}

\begin{equation}
\label{eq:efdm3}
    EFDMix(w,r)_{\tau_{i}} = w_{\tau_{i}} + (1 - d)r
    _{k_{i}} - (1 - d) \bigl \langle w_{\tau{i}} \bigr \rangle
\end{equation}

We utilize the instance-specific mixing weight denoted as $d$, which is obtained by sampling from a Beta distribution($c$, $c$), where $c \in (0, \infty)$ serves as a hyper-parameter. where $\bigl \langle \cdot \bigr \rangle$ represents the stop-gradient operation \cite{stopgrad}.

\textbf{Learnable mixup with style transfer}. For the purpose of adversarial domain augmentation, we employ feature perturbations. It is assumed that the perturbations, denoted as $e$, follow a multivariate Gaussian distribution $\mathcal{N}(\mu, \sigma^2)$. This allows for easy access to the uncertainty associated with the perturbations. The parameters of the Gaussian distribution $(\mu, \sigma)$ are learned using variational inference. 

The updated latent feature $h^{+}$ is obtained by adding the perturbations to the original feature $h$ with interpolated syle through EFDMix (r is obtained by shuffling $h$ along the batch dimension). This is denoted as: $h^{+} \leftarrow EFDMix(h, r) + e$, where $e$ is sampled from the Gaussian distribution $\mathcal{N}(\mu, \sigma^2)$. By employing this approach, we can create a series of feature augmentations in different training iterations. 

Our approach also involves blending $S$ and $S^{+}$ via Mixup \cite{mixup} to achieve intermediate domain interpolations. In particular, we utilize the uncertainty captured in the perturbations $(\mu, \sigma^2)$ to forecast adjustable parameters $(a, b)$, which guide the direction and intensity of domain interpolations.  

\vspace{-0.5cm}

\begin{equation}
\label{mixup}
    \begin{split}
    {h^{+}}^\prime &= \lambda \cdot EFDMix(h, r) + (1 - \lambda)h^{+}\\
    y^{+} &= \lambda y + (1 - \lambda)\tilde{y}
    \end{split}
\end{equation}

\noindent where: $\lambda \in Beta(a,b)$; $\tilde{y}$ is a smoothed version of $y$ by a chance of lottery $\tau$. Beta distribution \& lottery are computed by a fully connected layer following the Generator. These parameters integrate the uncertainty of generated domains.

\textbf{Adversarial domain augmentation}. In the latent space, we propose an iterative training procedure where two phases are alternated: a maximization phase where new data points are learned by computing the inner maximization problem and a minimization phase, where the model parameters are updated according to stochastic gradients of the loss evaluated on the adversarial examples generated from the maximization phase. The fundamental concept here involves iteratively acquiring "hard" data points from fictitious target distributions while retaining the essential semantic attributes of the initial data points via adversarial data augmentation \cite{erm}:

\vspace{-0.2cm}

\begin{equation}
\label{eq:adv}
    \max_{G} L
(M; S^+) - \beta \| \textbf{z} - \textbf{z}^{+} \|_2^{2}
\end{equation}

\noindent where: $L$ represents the cross-entropy loss and involves the creation of $S^{+}$ through the perturbations of $h^{+}$; the second term is the safety constraint that limits the greatest divergence between $S$ and $S^{+}$ in the embedding space. $z$ ($z^{+}$ when $G$ activated) denotes the output from the Projection head ($P$) and $\beta$ is a hyperparameter that controls the maximum divergence. The projection head part of our model transforms the convolutional features into a lower-dimensional feature space $Z$ suitable for contrastive learning. 
This is distinct from $h$ which denotes the outputs emerging from the conv. layers.

\vspace{0.15cm}

\noindent
\underline{\textbf{2.3. Learning objective}}:
%
%\subsection{Model objective}
%\label{sec:modelobj}
%
Here, our focus is on introducing the process of acquiring cross-domain invariant representations and producing effective domain augmentations $S^{+}$. Initially, we aim to maximize the conditional likelihood of the fictious domain $S^{+}$: $\log p(y^{+}|x, h^{+};M)$. That requires the true posterior $p(h^{+}|x;M,G)$ which is intractable. As a consequence, we use amortized variational inference with a tractable form of approximate posterior $q(h^{+}|x;M,G)$. The derived lower bound is expressed as follows:

\vspace{-0.3cm}

\begin{equation}
    L_{M,G} = \mathbf{E}_{q(h^{+}|{x};M,G})[\log\frac{p(y^{+}|x,h^{+};M)}{q(h^{+}|x;M,G)}]
\end{equation}

We use Monte-Carlo (MC) sampling to maximize the lower bound $L_{M,G}$ by:

\vspace{-0.4cm}

\begin{equation}
    \begin{split}
    L_{M,G} =\min_{M,G}\frac{1}{K}\sum_{k=1}^{K} \left[-\log p(y^{+}_{k}|x_k, h^{+}_{k};M) ) \right] + \\ KL\left[q(h^{+}|x;M,G) ||p(h^{+}|x;M,G)  \right]
    \end{split}
\end{equation}

\noindent where $h^{+}_{k} \sim q(h^{+}|x; M, G)$ and $K$ is the number of MC samples. For Kullback-Leibler divergence (KL) we let $q(h^{+}|x; M, G)$ approximate $p(h^{+}|x; M, G)$ through adversarial training on $G$ in Eq. \ref{eq:adv} so that the learned adversarial distribution is more flexible to approximate unseen domains. 

When the model is faced with a new domain, $T$, during testing, we compute its uncertainty with respect to the learned parameters of the task model $M$; $\sigma(S)$ through $\frac{\sigma(T) - \sigma((S)}{\sigma(S)}$ \cite{Qiao_2021_CVPR}. %Given that the generator $G$ consists of two convolutional layers, this score can be efficiently computed through single-pass data forwarding through $G$. 
In Section \ref{sec:exp}, we provide empirical evidence that our estimation aligns with traditional Bayesian methods \cite{bayesian}, while substantially reducing the time required, by an order of magnitude.

\vspace{-0.4cm}

\begin{equation}
    L_{NCE}(z_i, z_{i}^{+}) = \min_{M,G} - \log \frac{\exp(z_i \cdot z_{i}^{+})}{\sum_{j=1, j\neq i}^{2N} \exp(z_i \cdot z_j)}
\end{equation}

$L_{\text{NCE}}$ is the InfoNCE loss~\cite{contrastive} used for contrastive learning. In the minibatch $B$, $z_i$ and $z^{+}_{i}$ originate from the task model and perturbed task model (Generator G was activated) respectively. They share the same semantic content while originating from the source and augmented domain respectively. The utilization of $L_{\text{NCE}}$ will direct the learning process of $M$ towards acquiring domain-invariant representations by decreasing the distance between $z_i$ and $z^{+}_{i}$.

In summary, the loss function used to train the task model $M$ can be described as follows:

\vspace{-0.25cm}

\begin{equation}
    L_{total} = L_{M,G} + w_1 \cdot L_{NCE}(z_i, z_{i}^{+})
\end{equation}

\noindent
$w_1$: parameter controlling the importance of contrastive loss.

\vspace{0.15cm}

\noindent
\underline{\textbf{2.4. Datasets and Metrics}}:
%\subsection{Datasets and Setups}
%
%To assess the effectiveness of our approach, we employ two widely used domain generalization datasets: CIFAR-10-C \cite{cifarC} and PACS \cite{PACS}. 
\textbf{CIFA-10-C} \cite{cifarC}  is a robustness benchmark encompassing 19 diverse types of corruptions applied to the CIFAR-10 test set. These corruptions are classified into four primary categories: weather, blur, noise, and digital. 
\textbf{PACS} \cite{PACS} comprises four domains, each accompanied by a distinct number of images: Photo ($1670$ images), Art Painting (Art) ($2048$ images), Cartoon ($2344$ images), and Sketch ($3929$ images). The accuracy for each category and the average accuracy across all categories are presented as the main performance metrics. 

\noindent
\underline{\textbf{2.5. Implementation details}}:
We have tailored task-specific models and adopt distinct training strategies for the two datasets. We employ WideResNet \cite{wideresnet} (16-4) for CIFAR and AlexNet \cite{alexnet} for PACS, for fair comparison \cite{learningtodiversify, metacnn}. The Generator $G$ consists of two convolution layers (for $\mu$ and $\sigma$) and the projection head $P$ contains one fully connected layer.

\vspace{-0.3cm}

\section{Experimental Results}
\label{sec:exp}

\begin{table}[t]
\caption{The accuracy of single-source domain generalisation (\%) on CIFAR-10-C. Models are trained on CIFAR-10 and evaluated on the CIFAR-10-C.}
\label{CIFARC}
\centering
\scalebox{0.8}{
\begin{tabular}{|c||c|c|c|c|c|}
\hline
\textbf{Method}   & \textbf{Weather} & \textbf{Blur} & \textbf{Noise} & \textbf{Digital} & \textbf{Avg}\\
\hline
\hline
ERM \cite{erm}   & 67.28  & 56.73 & 30.02 & 62.30 & 54.08 \\
\hline
M-ADA \cite{mada}  & 75.54 & 63.76  & 54.21 & 65.10 & 64.65  \\
\hline
U-SDG \cite{uncertainty}   & 76.23  & 65.87 & 53.05 & 68.43 &  65.89\\
\hline
RandConv \cite{randconv}   & 76.87 & 55.36  &  75.19 & 77.51 & 71.23 \\
\hline
L2D \cite{learningtodiversify}   & 75.98 & 69.16  &  73.29 & 72.02 & 72.61\\
\hline
MetaCNN \cite{metacnn}   & 77.44 & 76.80  &  78.23 & 81.26 & 78.45\\
\hline
\hline
\begin{tabular}{@{}c@{}} \textbf{Ours} \\   \end{tabular}
   &\textbf{89.13} &  \textbf{82.94} &  \textbf{85.62} & \textbf{84.43} & \textbf{85.53} \\
\hline
\end{tabular}
}
\end{table}

\begin{table}[t]
\caption{The accuracy of single-source domain generalisation (\%) on PACS. Models are trained on photo and evaluated on the rest of the target domains.} %(i.e., art painting, cartoon, and sketch). Best performances are highlighted in bold.}
\label{PACS}
\centering
\scalebox{.8}{
\begin{tabular}{|c||c|c|c|c|}
\hline
\textbf{Method}   & \textbf{A} & \textbf{C}  & \textbf{S} & \textbf{Avg}\\
\hline
\hline
ERM \cite{erm}     & 54.43 & 42.74 & 42.02 & 46.39 \\
\hline
JiGen \cite{jigen}   & 54.98 & 42.62  & 40.62 & 46.07 \\
\hline
M-ADA \cite{mada}   & 58.96  & 44.09 & 49.96 & 51.00  \\
\hline

L2D \cite{learningtodiversify}   & 56.26 & 51.04   & 58.42 & 55.24\\
\hline
ALT \cite{altwacv}   & \textbf{68.50} & 43.50  &  53.30  & 55.10  \\
\hline
MetaCNN \cite{metacnn}   & 54.04 & \textbf{53.58}  &  \textbf{63.88} & 57.17\\
\hline
\hline
\begin{tabular}{@{}c@{}} \textbf{Ours} \\   \end{tabular}
   &59.30&  {50.66} &  {62.00} & \textbf{57.32} \\
\hline
\end{tabular}
}
\end{table}

\textbf{Comparison with the state-of-the-art} Tables \ref{CIFARC}, \ref{PACS} exhibit the evaluations of single domain generalisation on CIFAR-10-C and PACS, respectively. The results demonstrate that CUDGNet achieves the highest average accuracy compared to other methods. To be specific, as shown in Table \ref{CIFARC}, there are notable improvements of 11.69\%, 6.14\%, 7.39\%, 3.17\% and 7.08\% in weather, blur, noise, digital and average categories of CIFAR corruptions respectively. Table \ref{PACS} demonstrates that CUDGNet outperforms all previous methods in the domain of Art Painting except for \cite{altwacv} which has rather imbalanced results and achieves superior average performance. Sketch and Cartoon in contrast to Art painting have huge domain discrepancies compared to Photo (source domain), but still, our model achieves comparable results in these categories compared to the state-of-the-art. 
%These findings strongly suggest that the proposed CUDGNet framework enhances model generalization through the composition of safe and effective fictitious domain augmentations that help the model learn cross-domain invariant representations.

\vspace{0.1cm}

\begin{figure}[h!]
\centering
\includegraphics[width=.8\linewidth]{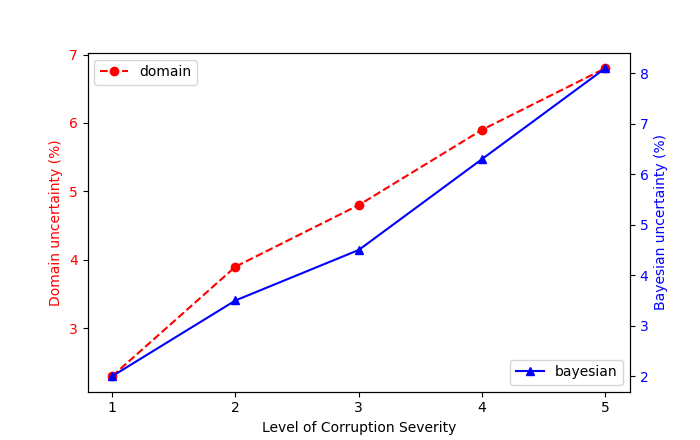}
\caption{Estimation of Uncertainty CIFAR-10-C. Our domain uncertainty prediction aligns with Bayesian uncertainty, while our approach is significantly faster.} 
\label{fig:uncertainty}
\end{figure}

\noindent
\textbf{Uncertainty estimation}. We compare the domain uncertainty score introduced in Section 2.3 with a more computationally intensive Bayesian approach \cite{bayesian}. The former calculates uncertainty through a single-pass forward operation ($~0.15$ ms/batch), whereas the latter relies on repeated sampling (30 times) to compute output variance ($~5.1$ ms/batch). In Figure \ref{fig:uncertainty}, we present the outcomes of uncertainty estimation for CIFAR-10-C. Evidently our estimation consistently aligns with Bayesian uncertainty estimation. This alignment underscores its high efficiency.

%\vspace{-0.6cm}
%($~0.15$ ms/batch), requiring only a single data forwarding pass, as opposed to the Bayesian method ($~5.1$ ms/batch).

\begin{table}[t]
\caption{Ablation study of the key components of CUDGNet}
\label{ablation}
\centering
\scalebox{0.8}{
\begin{tabular}{|c||c|c|c|c|c|}
\hline
\textbf{Method}   & \textbf{Weather} & \textbf{Blur} & \textbf{Noise} & \textbf{Digital} & \textbf{Avg}\\
\hline
\hline
Baseline  & 63.43  & 55.61 & 31.92 & 61.01 & 53.01  \\
\hline
+ G & 72.19  & 61.37 & 51.58 & 62.71 &  61.96\\
\hline
+ TC   & 79.22  & 75.44 & 71.46 & 75.14 & 75.31 \\
\hline
+ Style transfer    & 87.81 & 82.10  & 82.05  & 80.76 & 83.18 \\
\hline
+ Contrastive learning  & 89.13 & 82.94  & 85.62  & 84.43 & 85.53 \\

\hline
\end{tabular}
}
\end{table}

\noindent
\textbf{Ablation Study} %To investigate the contribution of each component, we performed an ablation study on the CIFAR-10-C dataset. 
As depicted in Table \ref{ablation} when incorporating the Transformation Component the model exhibits significant improvements over the baseline with adversarially augmented Generator. This underscores that slightly augmenting the diversity of the source domain and changing the image structure through fractals before the adversarial minmax optimization of the Generator is an effective strategy. Furthermore, the addition of style transfer results in a further performance boost of $7.87\%$. This suggests that this component empowers our model to generalise more effectively to novel domains. At last, with the integration of contrastive loss, we achieve a new state-of-the-art performance, boasting an average performance score of $85.53\%$. 

\section{CONCLUSION}

In this paper, we introduce a novel framework that enhances domain generalisation capabilities of the model and explainability through uncertainty estimation. Our comprehensive experiments and analysis have demonstrated the effectiveness of our method.

\bibliographystyle{IEEEbib}
\bibliography{strings}

\end{document}